\def\year{2018}\relax
\documentclass[letterpaper]{article} 
\usepackage{aaai18}  
\usepackage{times}  
\usepackage{helvet}  
\usepackage{courier}  
\usepackage{url}  
\usepackage{graphicx}  

\usepackage{latexsym}
\usepackage{epstopdf}
\usepackage{amsmath}
\usepackage{amssymb}
\usepackage{enumerate}
\usepackage[lined,boxed,commentsnumbered, ruled,linesnumbered]{algorithm2e}
\usepackage{stmaryrd}
\usepackage{mdwlist}
\usepackage{multirow}
\usepackage{color}

\begin{document}
\title{Table-to-Text: Describing Table Region with Natural Language}
\author{Junwei Bao$^\dag$\thanks{Contribution during internship at \mbox{Microsoft Research Asia}}, Duyu Tang$^\ddag$, Nan Duan$^\ddag$, Zhao Yan$^\S$, Yuanhua Lv$^\diamond$, Ming Zhou$^\ddag$, Tiejun Zhao$^\dag$\\
$^\dag$Harbin Institute of Technology, Harbin, China\ \ \
$^\ddag$Microsoft Research, Beijing, China\\
$^\S$Beihang University, Beijing, China\ \ \
$^\diamond$Microsoft AI and Research, Sunnyvale CA, USA\\
{\tt baojunwei001@gmail.com}\ \ \ {\tt \{dutang, nanduan, yuanhual, mingzhou\}@microsoft.com}\\
{\tt yanzhao@buaa.edu.cn} \ \ \ \ {\tt tjzhao@hit.edu.cn}
}


\maketitle

\begin{abstract}
  In this paper, we present a generative model to generate a natural language sentence describing a table region, e.g., a row.
  The model maps a row from a table to a continuous vector and then generates a natural language sentence by leveraging the semantics of a table. 
  To deal with rare words appearing in a table, we develop a flexible copying mechanism that selectively replicates contents from the table in the output sequence.
  Extensive experiments demonstrate the accuracy of the model and the power of the copying mechanism.
  On two synthetic datasets, W{\small IKI}B{\small IO} and S{\small IMPLE}Q{\small UESTIONS}, our model improves the current state-of-the-art BLEU-4 score from 34.70 to 40.26 and from 33.32 to 39.12, respectively.
  Furthermore, we introduce an open-domain dataset W{\small IKI}T{\small ABLE}T{\small EXT} including 13,318 explanatory sentences for 4,962 tables.
  Our model achieves a BLEU-4 score of 38.23, which outperforms template based and language model based approaches.
\end{abstract}

\section{Introduction}
A Table\footnote{\url{https://en.wikipedia.org/wiki/Table_(information)}} is a widely-used type of data source on the web, which has a formal structure and contains valuable information.
Understanding the meaning of a table and describing its content is an important problem in artificial intelligence, with potential applications like question answering, building conversational agents and supporting search engines.
\cite{pasupat2015compositional,sun2016table,yin2015neural,jauhartables,konstas2013global,li-EtAl:2016:EMNLP20162,yan-EtAl:2016:P16-11}.
In this paper, we focus on the task of table-to-text generation.
The goal is to automatically describe a table region (e.g., a row) with natural language.

The task of table-to-text could be used to support many applications, such as search engines and conversational agents.
On one hand, the task could be used to generate descriptive sentences for the structured tables on the web.
Current search engines could serve structured tables as answers by regarding the generated sentences as keys and tables as values.
On the other hand, tables could also be used as responses for conversational agents such as the intents of ticket booking and production comparison.
However, it is impractical for a conversational agent to read a table of multiple rows and columns on a smart-phone.
Table-to-text technology could transform the data into natural language sentences which could be sent back to users with utterances or voice via text-to-speech transformation.


The task of table-to-text generation has three challenges.
The first one is how to learn a good representation of a table.
A table has underlying structures such as attributes, cells and a caption.
Understanding the meaning of a table is the foundation of the following steps for table-to-text generation.
%
%
The second challenge is how to automatically generate a natural language sentence, which is not only fluent but also closely relevant to the meaning of the table.
The third challenge is how to effectively use the informative words from a table which are typically of low-frequency, such as name entities and numbers, to generate a sentence.

%
%

To address the aforementioned challenges, we introduce a neural network model that takes a row from a table and generates a natural language sentence describing that row.
The backbone of our approach is the encoder-decoder framework, which has been successfully applied in many tasks including machine translation \cite{kalchbrenner2013recurrent,sutskever2014sequence,bahdanau2014neural} and dialogue generation \cite{sordoni-EtAl:2015:NAACL-HLT}.
In the encoder part, the model leverages table structures to represent a row as a continuous vector.
In the decoder part, we develop a powerful copying mechanism that is capable of generating rare words from table cells, attributes and caption.
The entire model can be conventionally trained in an end-to-end fashion with back-propagation.
Furthermore, we introduce an open-domain dataset, W{\small IKI}T{\small ABLE}T{\small EXT}, including 13,318 explanatory sentences for 4,962 tables.


We conduct experiments on three datasets to verify the effectiveness of the proposed approach.
On W{\small IKI}T{\small ABLE}T{\small EXT}, our approach achieves a BLEU-4 score of 38.23, substantially better than template-based and neural language model based approaches by an order of magnitude.
Thorough model analysis shows that our copying mechanism not only dramatically boosts performance, but also has the ability to selectively replicate appropriate contents from a table to the output sentence.
On two synthetic datasets, W{\small IKI}B{\small IO} and S{\small IMPLE}Q{\small UESTIONS}, our approach improves the state-of-the-art BLEU-4 score from 34.7 to 40.26 and from 33.32 to 39.12, respectively.


This work makes the following contributions.
We present a neural network approach for table-to-text generation which effectively uses the structure of a table.
We introduce a powerful copying mechanism that is capable of generating rare words from a table.
We release an open-domain dataset W{\small IKI}T{\small ABLE}T{\small EXT}, and hope that it can offer opportunities to further research in this area.

\section{Task Formalization and Dataset}
We first formulate the task of table-to-text generation. Afterwards, we present the construction of an open-domain dataset W{\small IKI}T{\small ABLE}T{\small EXT}, which includes sentences that describe table regions.
\paragraph{Task Formalization.}
A table $T$ is defined as a tuple $T=\langle Attribute, Cell, Caption\rangle$, where
$Attribute=\{a_1,...,a_N\}$ includes $N$ attributes (column headers) of the table. 
$Cell=\{c_1^1,...,c_N^1,...,c_1^M,...,c_N^M\}$
includes $N*M$ cells of the table, where $N$ is the number of columns, $M$ is the number of rows, $c_i^j$ is the cell where the $i^{th}$ column and $j^{th}$ row interacts.  
$Caption$ is typically a natural language explanatory about the entire table.

\begin{figure}[htbp]
	\centering
    \includegraphics[width=3.28in]{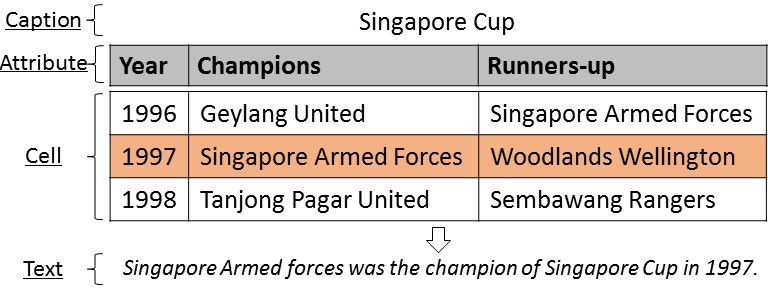}
	\makeatletter\def\@captype{figure}\makeatother \caption{An example of table-to-text generation.\label{example}}
\end{figure}

We formulate the task of table-to-text generation as follows.
Given a region of a table as input, the task is to output a natural language sentence to describe the selected table region.
In this work, we restrict the table region as a row, and leave a large region like multiple rows or entire table to future work.
Besides, information (cells) in the row can be \emph{selectively} used to generate a sentence.
Figure~\ref{example} gives an example that illustrates the task.
Given a selected row, which is highlighted in orange, the goal is to output a descriptive sentence ``\emph{Singapore Armed forces was the champion of Singapore Cup in 1997.}''.
In this case, only the information from the first two columns are used.

It is worth noting that, we deal with regular tables in this work and leave the irregular tables to future work.
We regard a table as a regular one if it does not contain merged attributes or merged cells, and the number of cells in each row is equal to the number of attributes.

\paragraph{W{\small IKI}T{\small ABLE}T{\small EXT}.\label{WTT-Dataset}}
We describe the construction of W{\small IKI}T{\small ABLE}T{\small EXT}.
We crawl tables from Wikipedia, and randomly select 5,000 regular tables, each of which has at least 3 rows and 2 columns.
For each table, we randomly select three rows, resulting in 15,000 rows that are further used for manual annotation.
Each annotator is given a selected row, the corresponding attributes and the caption.
We require that rows from the same table are labeled by different annotators.
If a table does not contain a caption, we use its page title instead.
Each annotator is asked to write a sentence to describe at least two cells from a table, but not required to cover every cell.
For example, the sentence in Figure~\ref{example} does not use the ``Runner-up'' column.
In addition, we also ask annotators not to search the meaning of a table from the web, as we would like to ensure that external knowledge is not used.
This makes the dataset more suitable for the real scenario.
To increase the diversity of the generated language, we assign different rows from the same table to different annotators.
If a row is hard to be described, we ask the annotator to write ``It's-hard-to-annotate''.
Finally, we get 13,318 row-text pairs. 
Statistics are given in Table~\ref{WTT-Table}.
We randomly split the entire dataset into training (10,000), development (1,318), and test (2,000) sets.

\begin{table}[htbp]
\centering
\small
\begin{tabular}{l|l}
    \hline
    \textbf{Type} & \textbf{Value} \\
      \hline
      Number of tables & 4,962      \\
      Number of sentences &  13,318     \\
      Avg \#sentences per table & 2.68\\
      Avg \#words per sentence & 13.91\\
      Avg / Min / Max \#words per caption & 3.55 / 1 / 14\\
      Avg / Min / Max \#cells per sentence & 3.13 / 2 / 9\\
      Avg / Min / Max \#columns per table & 4.02 / 2 / 10\\
      Avg / Min / Max \#rows per table & 7.95 / 3 / 19\\
      \hline
\end{tabular}
\caption{Statistics of W{\small IKI}T{\small ABLE}T{\small EXT}.\label{WTT-Table}}
\end{table}

To the best of our knowledge, W{\small IKI}T{\small ABLE}T{\small EXT} is the first open-domain dataset for table-to-text generation.
It differs from W{\small EATHER}G{\small OV}~\cite{liang2009learning} and R{\small OBO}C{\small UP}~\cite{chen2008learning} in that the schemes are not restricted to a specific domain, such as weather forecasting and RoboCup sportscasting.
We believe that W{\small IKI}T{\small ABLE}T{\small EXT} brings more challenges and might be more useful in real world applications.
We are aware that W{\small IKI}T{\small ABLE}Q{\small UESTIONS}~\cite{pasupat2015compositional} is a widely used dataset for table-based question answering task which takes a question and a table as input and outputs an answer.
However, we do not use this dataset in this work, because table-to-text generation task takes a row as input and outputs a sentence to describe the row, while
a portion of questions in W{\small IKI}T{\small ABLE}Q{\small UESTIONS} involve reasoning over multiple rows which does not satisfy the task constraint. 
We plan to handle multiple rows as input in future work.
Our task is also closely related to infobox-to-biography generation on W{\small IKI}B{\small IO}~\cite{lebret-grangier-auli:2016:EMNLP2016} and fact-to-question generation on S{\small IMPLE}Q{\small UESTIONS}~\cite{serban-EtAl:2016:P16-1}.
Both infoboxes and knowledge base facts can be viewed as special cases of tables.
Our task differs from them in that our input comes from a table with multiple rows and multiple columns.
Moreover, our dataset differs from~\cite{lebret-grangier-auli:2016:EMNLP2016} in that their dataset is restricted to biography domain, but our dataset is open-domain.
We differ from~\cite{serban-EtAl:2016:P16-1} in that their task generates questions, but our focus is to generate descriptive sentences.

\section{Background: Sequence-to-Sequence\label{S2S}}
Our approach is inspired by sequence-to-sequence (seq2seq) learning, which has been successfully applied in many language, speech and computer vision applications.
The main idea of seq2seq learning is that it first encodes the meaning of a source sequence into a continuous vector by an encoder, and then decodes the vector to a target sequence with a decoder.
In this section, we briefly introduce the neural network for seq2seq learning.
\paragraph{Encoder.}
The goal of the encoder component is to represent a variable-length source sequence $\mathbf{x}=\{x_{1},...,x_N\}$ as a fixed-length continuous vector.
The encoder can be implemented with various neural architectures such as convolutional neural network (CNN) \cite{meng2015encoding,gehring2016convolutional} and recurrent neural network (RNN) \cite{cho-EtAl:2014:EMNLP2014,sutskever2014sequence}.
Taking RNN as an example, it deals with a sequence by recursively transforming current words with the output vector in the previous step.
It is formulated as $h_{t}=f_{enc}(x_{t},h_{t-1})$
where $f_{enc}()$ is a nonlinear function, and $h_{t}$ is the hidden vector at time step $t$.
The last hidden vector $h_N$ is usually used as the representation of the input sequence $\mathbf{x}$.

\paragraph{Decoder.}
The decoder takes the output of the encoder, and outputs a target sequence $\mathbf{y}$.
Typically, the decoder is implemented with RNN, which generates a word $y_{t}$ at each time step $t\in \{1,2,...\}$ based on the representation of $\mathbf{x}$ and the previously predicted word sequence $y_{<t}=\{y_1,\ y_2,\ ...\ y_{t-1}\}$.
The process is formulated as follows,
\begin{equation}
p(\mathbf{y}|\mathbf{x})=\prod^{T}_{t=1}p(y_{t}|y_{<t},\mathbf{x})=\prod^{T}_{t=1}f_{dec}(y_{t-1},s_{t})\\ \nonumber
\end{equation}
where $f_{dec}()$ is a nonlinear function,
and $s_{t}=f_{hid}(y_{t-1},s_{t-1})$ is the hidden state of RNN at time step $t$.
Since standard RNN suffers from the problem of gradient vanishing~\cite{bengio1994learning}, existing studies usually use gated RNN units such as Long Short-Term Memory (LSTM)~\cite{hochreiter1997long} or Gated Recurrent Unit (GRU)~\cite{cho-EtAl:2014:EMNLP2014} 
for $f_{hid}()$.

\begin{figure*}[t]
	\centering
	\includegraphics[width=6.0in]{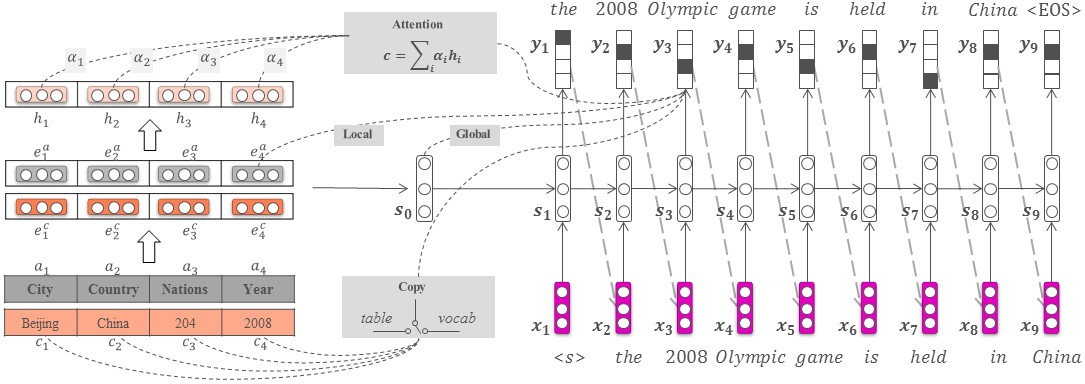}
	\caption{An illustration of the proposed approach for table-to-text generation.\label{ModelFigure}}
\end{figure*}

\section{Approach: Table-to-Sequence\label{Section_Model}}
We present a neural network approach for table-to-text generation in this section.
The approach is inspired by advances of seq2seq learning in neural machine translation, and we extend it by taking into account the table structure.
The intuition of our approach is to encode the meaning of a row from a table with an encoder, then generate a natural language sentence that describes the table with a decoder.
Figure~\ref{ModelFigure} shows an illustration of the model.

\subsection{Table Aware Encoder\label{T2S-Encoder}}
We implement a simple yet effective neural network to represent a row. We use information from a list of cells $\{c_1^j,...,c_N^j\}$ in the row, a list of attributes $\{a_1,...,a_N\}$ and a caption.
Specifically, we represent each cell $c_i^j$ as an embedding vector $e_i^{c} \in \mathbb{R}^{d_{c}}$, and represent each attribute $a_i$ as a continuous vector $e_i^{a} \in \mathbb{R}^{d_{a}}$.
As attributes and cells carry table information from different granularities, we use different embedding matrices to model them.
Afterwards, we get the representation $h_{i}=\tanh(W_e[e_i^{c};e_i^{a}] + b_e)$  of each column by concatenating $e_i^{c}$ and $e_i^{a}$, and then add a linear layer followed by element-wise nonlinear function.
As the majority of the captions have a small number of words, we regard a caption as a special cell and use a virtual attribute for it.
These column vectors $\{h_1, h_2, ..., h_N\}$ are used to represent the input row, and will be further used in the decoder part for calculating the attention probability.
In order to ensure that randomly exchanging two columns gets the same result, we simply use element-wise average over these column vectors, and use the result as the initial hidden state $s_{0}$ of the decoder.

\subsection{Table Aware Decoder\label{T2S-Decoder}}
We develop a table aware decoder that generates a sentence to describe a row. The backbone of the decoder is a GRU based RNN, which works in a sequential way and generates one word at each time step.
In order to take into account table structures and generate table-related descriptive sentences, we extend standard GRU-RNN in several ways, which will be detailed separately.

\paragraph{Attention Mechanism.}
When generating a word at a time step, the decoder should have the ability to selectively use important information from the table.
Furthermore, a descriptive sentence rarely uses the content of a cell more than once.
An example is given in Figure~\ref{ModelFigure}, where the decoder pays more attention to the cell ``2008'' at the second time step, and uses more information from the cell ``China'' at the 8-th time step.


Based on the considerations mentioned above, we adopt an attention mechanism, which assigns a probability/weight to each column at one time step.
Specifically, the attention weight of the $i$-th column at time step $t$ is calculated as
\begin{align*}
\alpha_{\langle t,i\rangle}=\frac{\exp{[z(s_{t},h_i,\sum^{N}_{j=1}\alpha_{\langle t-1,j\rangle}h_j)]}}{\sum^{H}_{i'=1}\exp{[z(s_{t},h_{i'},\sum^{N}_{j=1}\alpha_{\langle t-1,j\rangle}h_{j})]}} \nonumber
\end{align*}
where $h_i$ is the vector representation of the $i$-th column, $s_t$ is the hidden state of the standard GRU decoder, $z()$ is a non-linear function that computes the importance of $h_i$, which will be further normalized with a $softmax$ function.
The difference between this kind of attention mechanism with the well-known attention mechanisms~\cite{bahdanau2014neural,luong2015effective} is the incorporation of the attention results in previous time step.
Ideally, the model should have the ability to remember which columns have been used before, and not to use previously attended columns in the following time steps.
Similar techniques have been successfully applied in many tasks~\cite{chorowski2015attention,tu2016modeling,meng2016interactive,feng-EtAl:2016:COLING3}.

Afterwards, the attention weights are used to calculate a context vector $c_t$, which will further influence the hidden state $s_t$ at each time step $t$.
\begin{align*}
&c_{t}=\sum^N_{i=1}\alpha_{\langle t,i\rangle}h_i;
&s_{t}=\mathrm{GRU}(y_{t-1},s_{t-1},c_{t})\nonumber
\end{align*}

\paragraph{Global and Local Information.}
In order to increase the relevance between the generated sentence and the table, we further incorporate two types of table information into the decoder.
The first one is the \textbf{global} information about the entire table.
The intuition is that the decoder should use different implicit patterns to deal with tables about different topics, such as ``Olympic game'' and ``music album''.
As the output of the encoder captures information about the entire table, we set the global information as $s_0$ which is a part of the input of the target $softmax$ function.
%
Similarly, we use \textbf{local} information to remember the relation between the table and the generated word $y_{t-1}$ in the last time step.
Specifically, we regard the corresponding attribute of $y_{t-1}$ as the local information,
%
%
and feed its embedding $l_{t-1}$ as another part of the input of the target $softmax$.
We use a special symbol $<$\emph{unk\_a}$>$ to represent the attribute of $y_{t-1}$ if $y_{t-1}$ does not comes from the table.

\paragraph{Copying Mechanism.}
We observe that cells in a table typically include informative but low-frequency words, such as named entities and numbers.
These words are important to represent the meaning of a table, so that should be appropriately used to generate a meaningful sentence.
In the standard RNN decoder, a word is generated from a $softmax$ function, which is calculated over a word vocabulary of fixed length, such as the 20K most frequent words in the training data.
In this setting, rare words from table cannot be well covered in the vocabulary, so that they could not be generated in the predicted sequence.
Enlarging vocabulary size is not a principal way to solve this problem because it could not handle the rare words not seen in the training data.

Inspired by the recent success of copying mechanism in seq2seq learning \cite{gulcehre2016pointing,gu2016incorporating,yin2015neural,luong-EtAl:2015:ACL-IJCNLP,nallapati2016abstractive}, we develop a flexible copying mechanism that is capable of copying contents from table cells.
The intuition is that a word is generated either from target vocabulary via $softmax$ or from a table cell via the copying mechanism.
Specifically, we use a neural gate $g_{t}$ to trade-off between generating from target vocabulary and copying from the table cells.
\begin{equation}
g_{t}()=\sigma(W_{g}[W_{e}y_{t-1};s_{t};c_{t};s_0;l_{t-1}]+b_{g}) \nonumber
\end{equation}
where $\sigma$ is sigmoid function.
At time step $t$ , the probability of copying a word $\tilde{y}$ from table cells is $g_{t}(\tilde{y})\odot \alpha_{\langle t, id(\tilde{y})\rangle}$, where $id(\tilde{y})$ is the column index of $\tilde{y}$, and $\alpha_{\langle t, id(\tilde{y})\rangle}$ comes from the attention module.
The probability of generating a word $\tilde{y}$ from target vocabulary is $(1-g_{t}(\tilde{y}))\odot {\beta_{t}(\tilde{y})}$, where $\beta_{t}(\tilde{y})$ is calculated with the target $softmax$.
If a word $\tilde{y}$ is covered by both the target vocabulary and the table cells, its probability is the weighted sum of both sides.
\begin{equation}
p_{t}(\tilde{y})=g_{t}(\tilde{y})\odot \alpha_{\langle t, id(\tilde{y})\rangle}+(1-g_{t}(\tilde{y}))\odot {\beta_{t}(\tilde{y})} \nonumber
\end{equation}
%
The advantage of this copying mechanism is that
during training it does not have a preference for copying from table cells or generating from target vocabulary~\cite{gulcehre2016pointing}.
This property makes the model more flexible and could be conventionally trained and tested in the same way.

We name our model that copies contents from cells as \textbf{Table2Seq}.
We also go one step further and extend the copying mechanism to make it also copy words from attributes.
In the decoder part, we double the hidden length of encoder hidden states by adding the attribute embeddings, and regard the attention value of each attribute embedding as the evidence to copy from that attribute.
This attribute enhanced model is abbreviated as \textbf{Table2Seq++}.

\subsection{Training and Inference}
The model is trained in an end-to-end fashion using back-propagation under the objective
\begin{equation}
\mathcal{L}=-\frac{1}{|D|}\sum_{\langle \mathbf{x},\mathbf{\hat{y}} \rangle \in D}\sum^{T'}_{t=1}\log(p({\hat{y}}_{t}|{\hat{y}}_{<t},\mathbf{x})) \nonumber
\end{equation}
where $D$ is the training set.
In the inference process, we use beam search to generate the top-$K$ confident results where $K$ is the beam size.

\section{Experiment}
In this section, we describe experiment settings, and give the results and analysis on three datasets including W{\small IKI}T{\small ABLE}T{\small EXT}, W{\small IKI}B{\small IO} and S{\small IMPLE}Q{\small UESTIONS}.

\subsection{Implementation Details}

We use the same experiment setting for these three datasets.
We randomly initialize the parameters in our model with a Gaussian distribution, set the dimension of the word/attribute embedding as 300, and set the dimension of the decoder hidden state as 500.
We adopt Ada-delta~\cite{zeiler2012adadelta} to adapt the learning rate.
A dev set is used to half the learning rate when the performance on the dev set does not improve for 6 continuous epoches.
We update parameters in an end-to-end fashion using back-propagation.
In the inference process, we use beam search and set the beam size as 5.
We use BLEU-4~\cite{papineni2002bleu} score as our evaluation metric.
BLEU-4 score is widely used for natural language generation tasks, such as machine translation, question generation, and dialogue response generation.
Each instance in these datasets has only one reference sentence.

\subsection{Table-to-Text Generation with W{\small IKI}T{\small ABLE}T{\small EXT}}
\paragraph{Dataset.}
We conduct experiments on {W{\small IKI}T{\small ABLE}T{\small EXT}}.
We treat a caption as a special cell and use a virtual attribute ``caption" for it.
On average, a caption have 3.55 words.

\paragraph{Baselines.}
We implement a table conditional neural language model (\textbf{TC-NLM}) baseline, which is based on a recurrent neural network language model~\cite{mikolov2010recurrent}.
We feed the model with local and global factors to make it also consider table information.
We implement a \textbf{random-copying} baseline which does not use the copying mechanism but replaces the $<$\emph{unk}$>$ with a randomly selected cell from the table.
We also implement a \textbf{template}-based baseline.
During the training procedure, we construct a list of templates ranked by the frequency for each table scheme.
The templates are derived by replacing the cells appearing in the text with the corresponding attributes (slots).
In the inference process, we select the template with the highest frequency given a table scheme, and generate text by replacing the attributes (slots) with the corresponding values.

\paragraph{Results.}
\begin{table}[htpb]
	\small
	\centering
	\begin{tabular}{l|c|c}
		\hline
		Setting & Dev & Test\\ \hline\hline		
		TC-NLM & 5.31 & 5.79 \\
		Random-Copying & 11.11  & 12.01  \\
		Template &29.10& 28.62 \\ \hline
		Table2Seq                                    & 35.69 & {37.90} \\
		Table2Seq w/o Caption                              & 26.21 & 27.06 \\
		Table2Seq w/o Copying                               & 4.78  & 5.41  \\
		Table2Seq w/o Global                             & 34.82 & 36.68 \\
		Table2Seq w/o Local                              & 34.08 & 36.50 \\ \hline
		Table2Seq++                                    & 36.68 & \textbf{38.23} \\
		\hline
	\end{tabular}
	\caption{BLEU-4 scores on W{\small IKI}T{\small ABLE}T{\small EXT}.
		\label{Result_Table}}
\end{table}
Table~\ref{Result_Table} shows the experiment results of our model with different settings. 
The experiment results indicate that our method outperforms the baselines by an order of magnitude.
We remove each component of our model at a time, and show the influence of each part.
We find that copying mechanism is the most effective factor.
Removing the copying mechanism gets down the performance from 37.90 to 5.41.
Randomly copying values also performs bad.
It is reasonable because many words of the description come from table cells which are rare words and need to be copied.
\begin{table}[htbp]
  \small
  \centering
  \begin{tabular}{l}
  \hline
  \textbf{Ref}:~abe forsythe acted as ned kelly in movie ned in 2003 .\\
  \textbf{T2S}:~abe forsythe acted as ned kelly in the ned in 2003 .\\
  \textbf{T2S w/o Copying}:~$<$\emph{unk}$>$ played in $<$\emph{unk}$>$ in 2003 .\\ \hline
  \end{tabular}
  \caption{Examples of Table2Seq (T2S) with or without using copying mechanism. \textbf{Ref} is the reference sentence. 
  \label{ExamplesForNoCopy}}
\end{table}
Table~\ref{ExamplesForNoCopy} shows generated sentences by using copying mechanism or not.
Caption is the second important factor, which is usually an entity containing important information for generating the description.
On this dataset, the global and local factors do not have a big influence on the results.
In addition, contents of table attributes are also useful for describing a table.
We make a simple but effective adaptation and get an enhanced model, Table2Seq++,
which can copy from both the cells and attributes.
The results show that copying from both cells and attributes also brings improvements.
Table~\ref{copyAtt} are examples for copying attributes or not.
\begin{table}[htbp]
  \small
  \centering
  \begin{tabular}{l}
  \hline
  \textbf{Ref}:~the capacity of cyclone triple was triple .\\
  \textbf{T2S++}:~the capacity of cyclone triple was triple .\\
  \textbf{T2S}:~the $<$\emph{unk}$>$ of cyclone triple was triple .\\ \hline
  \end{tabular}
  \caption{Examples of Table2Seq++ (T2S++) and Table2Seq (T2S), where ``\emph{capacity}'' is an attribute. \textbf{Ref} is the reference sentence.
  \label{copyAtt}}
\end{table}

\begin{figure}[htbp]
	\centering
	\includegraphics[width=3.28in]{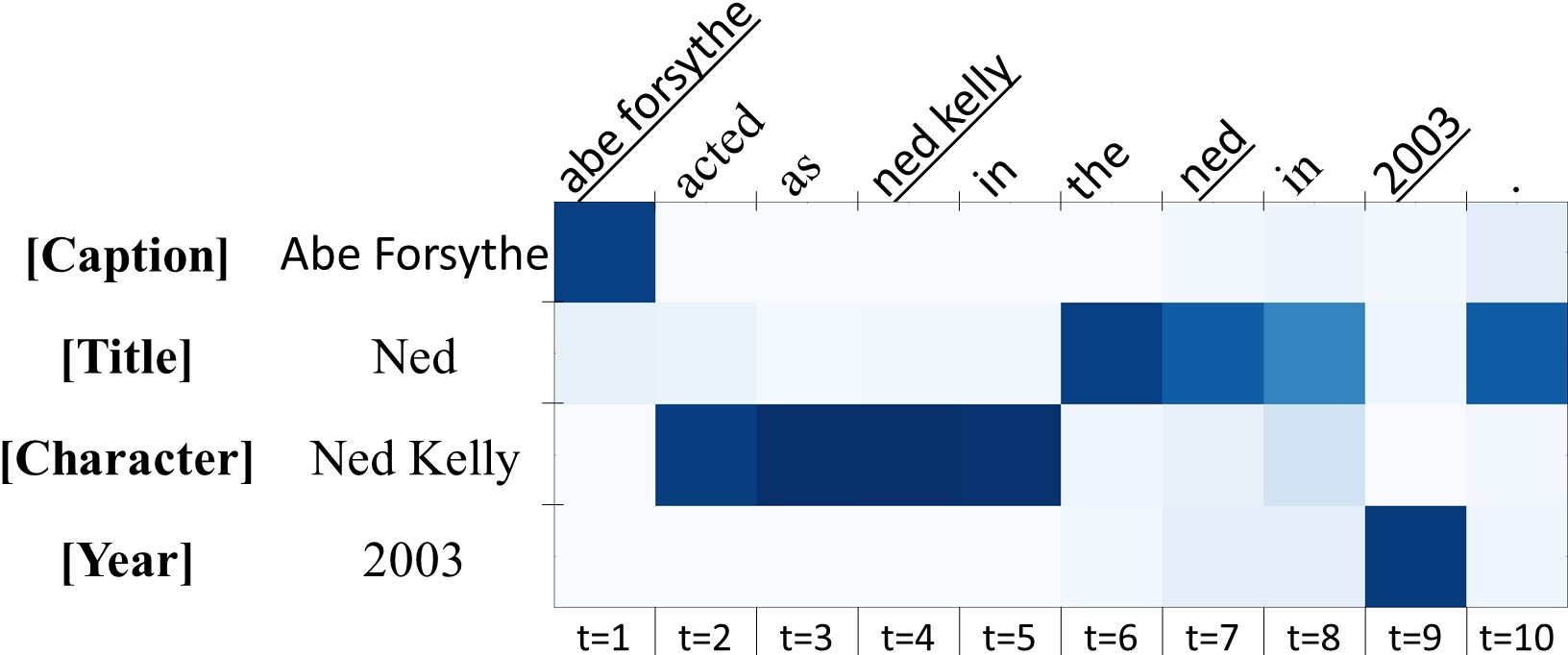}
	\caption{Visualization of attention probabilities.
		The vertical axis represents different table columns where words in square brackets are caption and attributes.
        The horizontal axis represents generated words at different time steps.
		Darker colour means higher attention probability.
		\label{Attention}}
\end{figure}

\begin{figure*}[htbp]
	\centering
	\begin{tabular}{c|l}
		\hline
		Group & Text \\ \hline\hline

		\multirow{2}{*}{(1)} & {\color{blue}{(\textbf{Reference})}}~ herman candries was the \textbf{chairman} of kv mechelen \textbf{in} 1977-82 . \\
		& {\color{red}{(\textbf{Generated})}} herman candries was the \textbf{president} of kv mechelen \textbf{during} 1977-82 .  \\ \hline

		\multirow{2}{*}{(2)} & {\color{blue}{(\textbf{Reference})}}~ stoycho mladenov got 21 goals in the game in 1978 .  \\
		& {\color{red}{(\textbf{Generated})}} stoycho mladenov got 21 goals in the \textbf{pfc beroe stara zagora} in 1978 .  \\ \hline

		\multirow{2}{*}{(3)} & {\color{blue}{(\textbf{Reference})}}~ c. narayanareddy won the \textbf{lifetime contribution award} in cinemaa awards in 2010 . \\
		& {\color{red}{(\textbf{Generated})}} c. narayanareddy was the champion of cinemaa awards in 2010 .\\ \hline
		
		\multirow{2}{*}{(4)} & {\color{blue}{(\textbf{Reference})}}~ the winner of \textbf{freedom cup} in 2011 was new zealand . \\
		& {\color{red}{(\textbf{Generated})}} new zealand was the winner of \textbf{1-1} in 2011 .\\ \hline
		
		\multirow{2}{*}{(5)} & {\color{blue}{(\textbf{Reference})}}~ nicholas rowe \textbf{acted} in movie girl on a cycle in 2003 .  \\
		& {\color{red}{(\textbf{Generated})}} nicholas rowe ( actor ) \textbf{released} girl on a cycle in 2003 .\\ \hline
		
		\multirow{2}{*}{(6)} & {\color{blue}{(\textbf{Reference})}}~ earl w. bascom won the grand marshal in 1984 .  \\
		& {\color{red}{(\textbf{Generated})}} earl w. bascom won \textbf{the grand marshal} in \textbf{the grand marshal} in 1984 .\\ \hline

	\end{tabular}
	\caption{Examples of the reference and generated sentences by {Table2Seq} model. \label{Examples_Table}}
\end{figure*}
\paragraph{Model Analysis.}
Since the experiment results indicate that the copying mechanism is the most effective component,
we further analyze the 
reason why it has such strong an impact on the results.
Specifically, as the attention probabilities are the main part of the copying mechanism, we visualize the attention distributions of an instance in Figure~\ref{Attention}, which shows probabilities for each table cell to be copied at each time step $t$.
In this example, the reference sentence is shown in Table~\ref{ExamplesForNoCopy},
which contains many table cells that are usually rare words.
Our model generates a meaningful and fluent sentence
which is almost the same as the reference.
The underlined words in the generated sentence are copied from the table cells. 
From the visualization we can see that, for each underlined word at time step $t=1,4,7,9$, the corresponding table cell it comes from has the highest probability of being copied.
This indicates that our model has the ability to properly copy rare words from the table contents.
In addition, we find that the learned gate has the ability of automatically deciding to copy from table contents or generate from target vocabulary.
For example, the word ``acted" at time step $t=2$ is generated even though the table cell ``Ned Kelly" has a high probability of being copied.
\begin{figure}[htbp]
	\centering
	\includegraphics[width=3.28in]{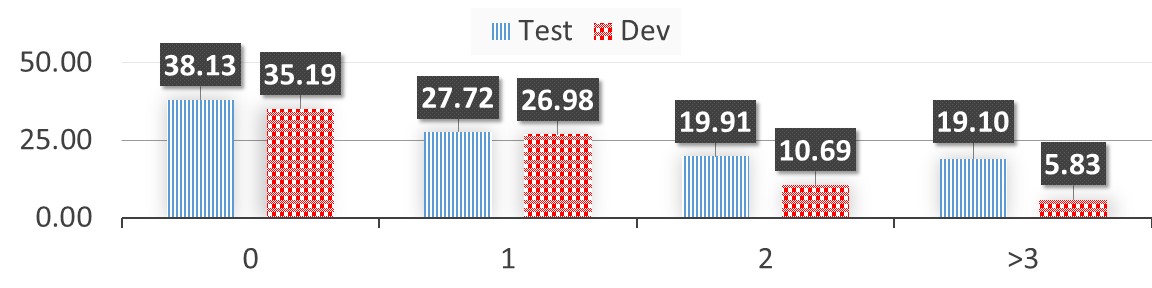}
	\caption{BLEU-4 scores on each portion of the dev/test set.
		Horizontal axis represents the number of unseen attributes in each table of that portion. Vertical axis represents the BLEU-4 score.
		\label{attributeAnalysis}}
\end{figure}

In the test process, an attribute that is not seen in the training corpus is represented as a special $<$\emph{unk\_a}$>$. In this part, we conduct experiments to explore the effects of $<$\emph{unk\_a}$>$ to the performance of our model.
This reveals the ability of our model to deal with the unseen attributes.
Therefore,
we respectively split the dev and test sets into four portions based on the number (0, 1, 2, or $>$3) of unseen attributes in a table.
Each portion of the dev (test) set contains 97.34\%, 2.20\%, 0.38\% and 0.08\% (97.4\%, 2.00\%, 0.50\% and 0.10\%) instances.
From Figure~\ref{attributeAnalysis}, we find that the BLEU-4 score decreases with the increasing of the number of unseen attributes in a table.
How to effectively deal with $<$\emph{unk\_a}$>$ is left for future work.

\paragraph{Case Study and Discussion.}
We sample 100 examples from the validation set and perform a case study to analyze the advantages of our approach and present challenges of the task.
Different groups of examples are shown in Figure~\ref{Examples_Table}.

From the results in groups 1, 2 and 3, we find that our model could generate fluent and table related sentences.
In group 1, the model generates a sentence with the same meaning as the reference, but uses different expressions. 
In groups 2 and 3, our model generates sentences with more or less information than the references as we do not restrict which columns are used. 
For example, the table cell ``pfc before stara zagora" in group 2 as additional information is also generated, while table cell ``lifetime contribution award" in group 3 is skipped.
Although the generated sentences from groups 1 to 3 are meaningful and fluent, they may lower the BLEU-4 score. 

From groups 4 to 6, we find that there still exist challenges in table-to-text generation task.
Sometimes our model copies a cell from a wrong column as the meaning of the columns are not well understood.
For example in group 4, the model copies ``1-1" in column ``Matched\_Won" but not ``freedom cup" in the ``Caption".
Similarly, some generated sentences use wrong words to describe the relation of two table cells.
For example in group 5, the word ``release" does not correctly describe the relation between cell ``nicholas rowe ( actor )" and ``girl on a cycle".
Incorporating more background knowledge, such as a knowledge base or table surrounding text, is a potential solution to this problem.
Even though we use recurrent attention to avoid over generating duplicate words, there still exist a few instances with the problem of duplication, e.g., group 6.

\subsection{Infobox-to-Biography Generation with W{\small IKI}B{\small IO}}
\paragraph{Dataset.}
W{\small IKI}B{\small IO} is introduced by~\cite{lebret-grangier-auli:2016:EMNLP2016} for generating biography to describe an infobox.
An infobox can be viewed as a table with one row.
They extract the first sentence of each biography article as reference.
On average, each reference sentence has 26.1 words.
The corpus contains 728,321 instances, which has been divided into three sub-parts to provide 582,659 for training, 72,831 for validation and 72,831 for testing.
In our experiments, we only use 2,000 instances for validation.
A characteristic of this dataset is that, each cell of an infobox often contains multiple words.
Therefore,
as suggested by \cite{lebret-grangier-auli:2016:EMNLP2016},
we split one cell into multiple cells where each new cell contains only one word.
\begin{table}[htpb]
	\small
	\centering
	\begin{tabular}{l|c|c}
		\hline
		Setting & Dev & Test\\ \hline \hline
		KN & - & 2.21 \\
		TC-NLM & - & 4.17 \\
		Template KN & - & 19.80 \\
		Table NLM & - & 34.70 \\ \hline
		Table2Seq                                    & 40.33 & \textbf{40.26}\\ 
		Table2Seq w/o Copying                               & 37.28 & 36.88 \\
		Table2Seq w/o Global                             & 40.69 & 40.11 \\
		Table2Seq w/o Local                              &40.43 &40.03\\
		\hline
	\end{tabular}
	\caption{BLEU-4 scores on W{\small IKI}B{\small IO}.
	\label{Result_WikiBio}}
\end{table}
\paragraph{Results and Analysis.}
Table~\ref{Result_WikiBio} shows the experiment results on W{\small IKI}B{\small IO}.
We choose methods introduced by~\cite{lebret-grangier-auli:2016:EMNLP2016} as our baselines, where the ``Table NLM" method is considered state-of-the-arts. 
The results show that our Table2Seq method outperforms the baselines.
As we observe that attributes are usually not expressed in a biography, we do not use Table2Seq++.
We remove each factor at a time and find that the results have the same trend as previous results on W{\small IKI}T{\small ABLE}T{\small EXT}.
On this dataset, removing the copying mechanism does not dramatically decrease performance.
We think this is because the average sentence length on W{\small IKI}B{\small IO} is much longer than W{\small IKI}T{\small ABLE}T{\small EXT} ($26.1>13.9$), so that the rare words have relatively little influence on the results.
In addition, our model without a copying mechanism performs better than ``Table NLM".
We believe the gain comes from the GRU-based decoder with recurrent attention.

\subsection{Fact-to-Question Generation with S{\small IMPLE}Q{\small UESTIONS}}
\paragraph{Dataset.}
We follow~\cite{serban-EtAl:2016:P16-1} to generate questions from knowledge base (KB) facts on S{\small IMPLE}Q{\small UESTIONS}~\cite{bordes2015large}.
A fact in KB is a triple containing a subject, a predicate and an object.
We regard a fact as a table with two attributes and a row with two cells.
S{\small IMPLE}Q{\small UESTIONS} contains 108,442 fact-question-answer tuples.
The dataset is split into three parts: 75,910 for training, 10,845 for validation, and 20,687 for test.

\begin{table}[htpb]
	\small
	\centering
	\begin{tabular}{l|c|c}
		\hline
		Setting & Dev & Test\\ \hline\hline
		Template& - & 31.36 \\
SP Triples & - & 33.27 \\
MP Triples & - & 32.76 \\
SP Triples TransE++ & - & 33.32 \\
MP Triples TransE++ & - & 33.28 \\ \hline
		Table2Seq                                    & 40.16 & {38.85}\\
		Table2Seq w/o Copying                               & 14.69 & 14.70 \\
		Table2Seq w/o Global                             & 40.34 & \textbf{39.12} \\
		Table2Seq w/o Local                              & 39.11 & 38.32 \\
		\hline
	\end{tabular}
	\caption{BLEU-4 scores on S{\small IMPLE}Q{\small UESTIONS}.
		\label{Result_SimpQ}}
\end{table}
\paragraph{Results and Analysis.}
We use the methods for fact-to-question generation reported by~\cite{serban-EtAl:2016:P16-1} on S{\small IMPLE}Q{\small UESTIONS} as our baselines.
The baselines first generate a question template with a place holder, then replace the place holder with the subject as post processing.
Different from their methods, our model automatically learns where to copy from.
Table~\ref{Result_SimpQ} shows the experiment results of the baselines and our model which is exactly the same as the one used in previous experiments.
The results indicate that our method outperforms the baselines by a wide margin. As predicates do not appear in questions, we do not evaluate the results on Table2Seq++.
Furthermore, the results verify that our model has the ability to generate natural language \emph{questions} from KB facts if it has rich training datasets.

\section{Related Work}
Our work relates to existing studies that generate natural language sentence from structured data~\cite{barzilay2005collective,reiter2005choosing,angeli2010simple,konstas2012concept,androutsopoulos2013generating}.
Recently, \cite{wiseman2017challenges} conduct experiments on generating long text from tables and show challenges in data-to-text generation task.
This work differs from existing studies in that the structured data in this work is a web table.
In \cite{angeli2010simple} and \cite{mei2015talk}, the structured data is a scenario consisting of a set of database records.
In \cite{lebret-grangier-auli:2016:EMNLP2016}, the structured data is an infobox consisting of fields and field values.
In \cite{serban-EtAl:2016:P16-1}, the structured data is a fact consisting of a subject, predicate, and object from Freebase.
This work differs from existing neural network based methods in that our approach is an encoder-decoder approach with attention and copying mechanisms.
\cite{mei2015talk} and \cite{serban-EtAl:2016:P16-1} also adopt neural encoder-decoder approaches 
, while they do not leverage copying mechanism to replicate source contents in output.
The approach of \cite{lebret-grangier-auli:2016:EMNLP2016} is based on a conditional language model to generate biographies.

\section{Conclusion}
We present an encoder-decoder approach for generating a sentence to describe a table row.
The model extents the encoder-decoder framework by integrating the semantics of a table.
A flexible copying mechanism is developed to replicate contents from cells, attributes, and captions in the output sentence.
We release an open-domain dataset W{\small IKI}T{\small ABLE}T{\small EXT} for table-to-text generation, and conduct extensive experiments on W{\small IKI}T{\small ABLE}T{\small EXT} and two synthetic datasets.
Experiment results demonstrate the accuracy of our approach and the power of our copying mechanism.

\section{Acknowledgments}
The research work is supported by the National Key Research and Development Program of China under Grant No.2017YFB1002102. We thank Zhirui Zhang for the valuable discussion on building the model. We also thank all the reviewers for their valuable comments.

\fontsize{9.5pt}{10.4pt} \selectfont \bibliography{reference}
\bibliographystyle{aaai}
\end{document}